\documentclass{article}

\usepackage{arxiv}

\usepackage[utf8]{inputenc}
\usepackage[T1]{fontenc}
\usepackage{hyperref}
\usepackage{url}
\usepackage{booktabs}
\usepackage{amsmath}
\usepackage{amssymb}
\usepackage{amsfonts}
\usepackage{nicefrac}
\usepackage{microtype}
\usepackage{float}
\usepackage{graphicx}
\usepackage{subcaption}
\usepackage{multirow}
\usepackage{booktabs}
\usepackage{threeparttable}
\usepackage{nicefrac}
\usepackage[justification=centering]{caption}

\title{Short-Term Traffic Flow Prediction Using Variational LSTM Networks}

\author{
	Mehrdad Farahani\\
	Department of Computer Engineering\\
	Islamic Azad University North Tehran Branch\\
	Tehran, Iran \\
	\texttt{m3hrdadfi@gmail.com} \\
	\And
	 Marzieh Farahani\\
	Department of Computing Science\\
	Umeå University\\
	Umeå, Sweden\\
  	\texttt{mafa2431@student.umu.se} \\
   	\And
	Mohammad Manthouri\\
	Department of Electrical and Electronic Engineering\\
	Shahed Univerisity\\
	Tehran, Iran\\
	\texttt{mmanthouri@shahed.ac.ir} \\
	\And
	Okyay Kaynak\\
 	Department of Electrical and Electronic Engineering\\
	Bogazici University\\
	Istanbul, Turkey\\
	\texttt{okyay.kaynak@boun.edu.tr}\\
}

\date{}

\begin{document}
\maketitle

\begin{abstract}
Traffic flow characteristics are one of the most critical decision-making and traffic policing factors in a region. Awareness of the predicted status of the traffic flow has prime importance in traffic management and traffic information divisions. The purpose of this research is to suggest a forecasting model for traffic flow by using deep learning techniques based on historical data in the Intelligent Transportation Systems area. The historical data collected from the Caltrans Performance Measurement Systems (PeMS) for six months in 2019. 
The proposed prediction model is a Variational Long Short-Term Memory Encoder in brief VLSTM-E try to estimate the flow accurately in contrast to other conventional methods. VLSTM-E can provide more reliable short-term traffic flow by considering the distribution and missing values.
\end{abstract}


\section{Introduction}
Urban life has undergone many changes in the development of local communities. This transport transformation and traffic congestion lead to road-clogging, slower speeds, longer trip times, and increased vehicular queuing in most of the urban and suburban passages in the world. This issue will be the trigger of abundant problems such as air pollution and noise pollution and in total, has a massive role in quality reductions. Therefore, governors recognize intelligent traffic flow control systems as a priority plan for their countries. The traffic flow forecasting is a crucial step for obtaining time optimizers in the public traffic adaptive control system.  

Traffic flow prediction is a significant issue for both transport management from one side and drivers and ordinary people on the other side. These methods help managers to recognize heavy traffics in the countrysides. Using some predefined paradigms and protocols can avoid the incidence of long traffic jams. On the other hand, drivers and ordinary people can also make a better decision based on that prediction and contributing to decreasing traffic levels. Therefore, predicting traffic flow characteristics in a geographical area is one of the most critical decision-making and policymakers that have a significant effect on urban traffic management. Mainly traffic flow prediction divided into three categories \cite{Hou2016RepeatabilityAS}.

\begin{itemize}
    \item Short-term forecasting (the interval is 5 minutes to 30 minutes)
    \item Medium-term forecasting (a time interval of 30 minutes to several hours)
    \item Long-term forecasting (ranges of one day to several days)
\end{itemize}

The ultimate goal in this domain is to evaluate the traffic flow prediction with the historical traffic data in a particular region before it happens. However, unpredictable disturbances, including internal-events in transportation ways (such as an accident, falling part of the route) and unexpected external-events (such as a flood, storm) make long-term forecasting inaccurate enough. While medium-term or short-term forecasting can be reliable if they correctly setup. 

In this research, the short-term case takes into consideration. The hybrid deep learning method predicts the flow based on a complex generative model from the data, which can recognize the spatial and temporal correlation within the sequence of traffic flows in a particular range. Furthermore, in the following, the recommended model compares to other state-of-the-art models.

The contribution of this paper can be summarized as follows:
\begin{itemize}
    \item Presenting a novel hybrid deep learning model based on a Variational Long Short-Term Memory Encoder (VLSTM-E)
    \item The proposed model is considering the distribution of data to forecast short-term traffic flow
    \item Take into consideration the missing data, which occurred by sensors failure by the distributed data
\end{itemize}

The paper is segmented as follows; the next section gives a brief description of terminologies, challenges, and other methods of short-term traffic forecasting research concerning several neural network techniques. In section 3, the background of the model is introduced. Then, In section 4, the suggested model is presented. The dataset is denoted in section 5, and the results, and performance evaluation are presented in section 6. Finally, conclusions and future research are stated in section 7.

\section{Related Works}
Traffic flow forecasting is one of the most useful tools in intelligent transportation systems (ITS). It allows the system to be in a control automatic operation state and anticipates the events before they occur. It can be able to predict and assess the states and prepare itself for logical decision-making at the machine level, and based on human-made protocols can manage the condition \cite{Oh2015UrbanTF}. Meanwhile, the short-term prediction of the traffic flow is more critical than the other two before categories in the field of intelligent transportation systems, in which many research and development are done in both academically and operationally \cite{Oh2015UrbanTF}. A great deal of research on the short-term forecasting model can be classified into two main categories:

\begin{itemize}
    \item Parametric, Including methods such as state-space methods \cite{Stathopoulos2003}, Kalman filter methods \cite{Zhou2019}, spectra analysis methods \cite{Zhang2014}, statistical techniques \cite{Krblek2019}, ARIMA, ARIMAX, and SARIMA models \cite{Luo2018,Hou2019,Ihueze2018}, and Markov model \cite{Zhu2016,Zhang2017}.
    
    \item Nonparametric, In these models, with non-linear backgrounds, we are trying to find the model that has the most receptive learning features. Many research has gotten lots of remarkable results with this insight, such as non-parametric regression techniques \cite{Huang2007,Agarwal2016,Apronti2016}, k-nearest neighbor models \cite{Cai2016}, fuzzy techniques \cite{Sharma2016,Guo2018,Chen2018}, neural networks \cite{Goves2016,Raj2016,Li2017,Sharma2018,Wang2019_2}, and support vector machine \cite{Cheng2017,Sun2015,Xiao2018}.
\end{itemize}

The spatial-temporal real-time information by traffic sensors around the country is one of the signs of technological advancement that brings up valuable facilities for the transportation systems of the country. The information provides a massive amount of patterns and paradigms of terrestrial transport in a geographic location. Moreover, the direct and indirect effects of that information present the foundation for the application of deep learning networks. Deep learning is a section of machine learning that grants short-term forecasts of traffic flows to find latent dependence relationships in a set of patterns with high dimensions of explanatory variables. This model tries to detect extreme disturbances in the traffic flow within a pool of latent relations providing by real-time sensors \cite{Polson2017,Wu2018}. Nevertheless, there is no clue that which types of deep learning models are the most appropriate model for forecasting traffic flows. All of these models are trying to find a part of these latent relations by presenting a different structure.

For example, the Stacked Autoencoders model was introduced by considering time and space correlation, was able to learn the general characteristics of the traffic flow \cite{Lv2014}. Another model that was able to achieve better performance is the Long Short-Term Memory (LSTM) and Gated Recurrent Unit (GRU) networks \cite{Fu2016}. These models provided a solution for gaining better results with an increase in the length of the sequences of information. It is necessary to take into account the effects of time before, and after more on each day. The performance of these models is significantly downed due to the accumulation of errors. The LSTM+ model in \cite{Yang2019} made it possible to achieve better performance considering these effects. 

In addition to predicting traffic flow behavior, which is one the importance of the traffic flow prediction,  traffic sensors are usually controlling manually, so these collections of data from sensors accompany with various lengths, irregular sampling, and missing data. These dissonances make this prediction complicated.  To solve this challenge, the researcher proposed a model base on Long Short-Term Memory in \cite{Tian2018}. Also, Convolutional Neural Network models, which showed their abilities to resolve image issues, are used in this domain so that they could provide excellent results in prediction the traffic flow \cite{Wang2019}.

\section{Background}
Since the central core of the proposed mode divided into two parts, variational and Long Short-Term Memory (LSTM). In the following, each section introduced in detail.

\subsection{Long Short-Term Memory}\label{lstm_section}
\begin{figure}[!htbp]
\centering
\centerline{\includegraphics[width=0.5\textwidth]{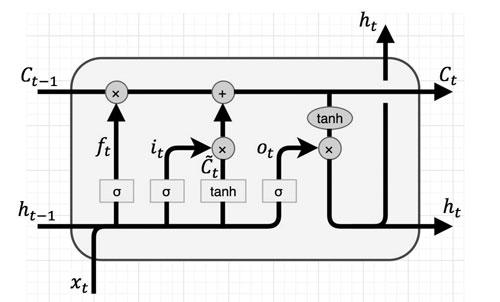}}
\caption{Long short-term memory cell.}
\label{fig:lstm}
\end{figure}

Long short-term memory (LSTM), as shown in Fig (\ref{fig:lstm}), proposed by \cite{Hochreiter1997}, is a recursive neural network architecture that is capable of learning long-term dependencies. This model has been developed to deal with vanishing gradient problems and considered a deep neural network architecture over time. The main component of the Long short-term memory layer is the memory cell.  

A memory cell consists of four main elements: an input gate, a neuron with reconnection, a forget gate, and an output gate. The following equations show step by step operation of a layer of memory cells for input time series as $X = ( x_{1}, x_{2}, x_{3}, ..., x_{n} )$, hidden states memory cells $H = ( h_{1}, h_{2}, h_{3}, ..., h_{n} )$.

\begin{align}
i_{t} &= \sigma\big(x_{t}U^{i}+h_{t-1}W^{i}\big) \\
f_{t} &= \sigma\big(x_{t}U^{f}+h_{t-1}W^{f}\big) \\
o_{t} &= \sigma\big(x_{t}U^{o}+h_{t-1}W^{o}\big) \\
\tilde{C}_{t} &= \tanh\big(x_{t}U^{g}+h_{t-1}W^{g}\big) \\
C_{t} &= \sigma\big(f_{t}\ast C_{t-1}+i_{t}\ast\tilde{C}_{t}\big) \\
h_{t} &= \tanh(C_{t})\ast o_{t}
\end{align}

The $*$ sign in this calculation considered as element-wise multiplication, and by refusing the bias terms, it can be shown how the hidden layer calculated at a time $h_{t}$. In the calculations above:

\begin{itemize}
    \item $i, f, o$ are called the input, forget and output gates, respectively.
    \item $W^{i}, W^{f}, W^{o}$ the weights connect the recurrence layer at $t-1$ to the hidden layer at time $t$.
    \item $U^{i}, U^{f}, U^{o}$ weights that connect the hidden layer at time $t-1$ to the recursive layer at time $t$.
\end{itemize}

At the end of the weighted non-linear calculation in the gates section, the output enters int a sigmoid activation function so that it can simulate the gating concept since the sigmoid activation function as shown in Eq (\ref{eq:sigmoid_activation}) with a range from 0 to 1 can provide a gateway as an open or closed concept

\begin{align}\label{eq:sigmoid_activation}
\sigma_{x} &=  \frac{1}{1 +  e^{x} } 
\end{align}

In Long Short-Term Memory networks, the objective function can be different depending on the structure of the problem, which cross-entropy, softmax, and l quadratic can be called accessible functions.

\subsection{Variational Autoencoders}
\begin{figure}[!htbp]
\centering
\centerline{\includegraphics[width=1.0\textwidth]{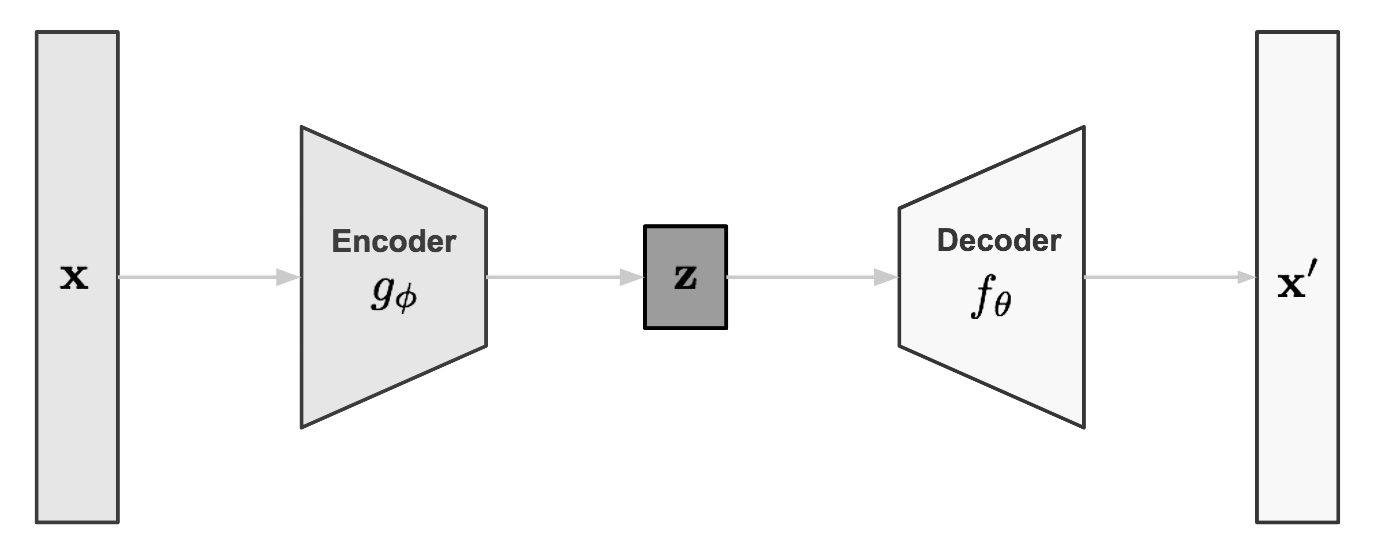}}
\caption{Autoencoder model architecture.}
\label{fig:ae_arch}
\end{figure}

Before paying attention to the variational part, it is necessary to get acquainted with the concept of an Autoencoder \cite{Schmidhuber2015DeepLI}. The Autoencoder network is a bipartite neural network that teaches the network to compress the information by forcing an encoder network to the output in that case to a low dimensional representation $z$, which is then consumed by a decoder network to output the original data as shown in (\ref{fig:ae_arch}).

\begin{figure}[!htbp]
\centering
\centerline{\includegraphics[width=1.0\textwidth]{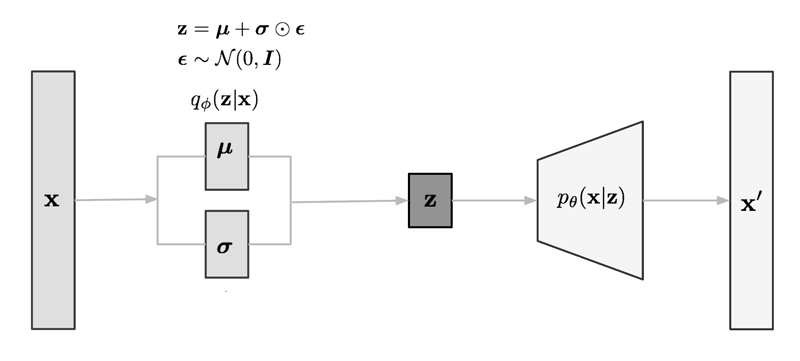}}
\caption{Variational Autoencoder model with the multivariate Gaussian assumption}
\label{fig:vae_gaussian}
\end{figure}

However, concerning the variational part \cite{Kingma2014AutoEncodingVB}, we must say that the goal is to achieve a model in which reproduction is not dependent only on data. Variational Autoencoder tries to decode data from some known probability distribution, in this case, Gaussian distribution that comes from encoding part to produce reasonable outputs even if they are not encoding actual data as shown in Fig (\ref{fig:vae_gaussian}).

\begin{figure}[!htbp]
\centering
\centerline{\includegraphics[width=0.7\textwidth]{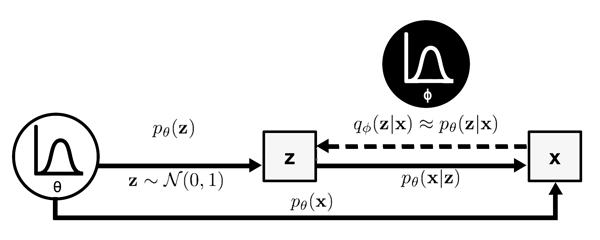}}
\caption{The graphical model of Variational Autoencoder. Solid lines denote the generative distribution $p_\theta(\mathbf{z})$, and dashed lines denote the distribution $q_\phi(\mathbf{z}\vert\mathbf{x})$ to approximate the intractable posterior $p_\theta(\mathbf{z}\vert\mathbf{x})$.}
\label{fig:vae_graphical_model}
\end{figure}

Suppose ${ \mathbf{x} = x^{(1)}, x^{(2)}, x^{(3)}, ..., x^{(N)} }$ be a set of observed variables and ${ \mathbf{z} = z^{(1)}, z^{(2)}, z^{(3)}, ..., z^{(M)} }$  be a set of hidden variables with joint distribution $p(Z, X)$. Label this distribution as $p_{\theta}$ which parameterized by $\theta$. To generate a sample that looks like a real data point $x^{(i)}$ as shown in Fig (\ref{fig:vae_graphical_model}).  

Then the inference issue is to calculate the conditional distribution of hidden variables given the observations, that is, $p_{\theta}(\mathbf{z}|\mathbf{x})$ which can write as shown in Eq (\ref{eq:posterior_dist}). 

\begin{align}\label{eq:posterior_dist}
p_{\theta}(\mathbf{z}|\mathbf{x}) &=  \frac{p_{\theta}(\mathbf{z}, \mathbf{x})}{p_{\theta}(\mathbf{x})}  \\
p_{\theta}(\mathbf{x}) &= \int p_{\theta}(\mathbf{x}\vert\mathbf{z}) p_{\theta}(\mathbf{z}) d\mathbf{z} \notag
\end{align}

Unfortunately, computing $p_{\theta}(\mathbf{x})$ is quite difficult because it is very expensive to check all the possible values of $\mathbf{z}$ and sum them up. So, to solve this issue, approximate $p_{\theta}(\mathbf{z}|\mathbf{x})$ by another distibution $q_{\phi}(\mathbf{z}|\mathbf{x})$ then can perform approximate inference of the intractable distribution. In order to ensure that  $q_{\phi}(\mathbf{z}|\mathbf{x})$ and $p_{\theta}(\mathbf{z}|\mathbf{x})$ were similar to each other, we could minimize the KL divergence between these two distributions, as shown in Eq (\ref{eq:kl_divergence}).

\begin{align}
\label{eq:kl_divergence}
& D_\text{KL}( q_\phi(\mathbf{z}\vert\mathbf{x}) \| p_\theta(\mathbf{z}\vert\mathbf{x}) ) \\
&=\int q_\phi(\mathbf{z} \vert \mathbf{x}) \log\frac{q_\phi(\mathbf{z} \vert \mathbf{x})}{p_\theta(\mathbf{z} \vert \mathbf{x})} d\mathbf{z} \notag\\
&=\int q_\phi(\mathbf{z} \vert \mathbf{x}) \log\frac{q_\phi(\mathbf{z} \vert \mathbf{x})p_\theta(\mathbf{x})}{p_\theta(\mathbf{z}, \mathbf{x})} d\mathbf{z} \notag\\
&=\log p_\theta(\mathbf{x}) + D_\text{KL}(q_\phi(\mathbf{z}\vert\mathbf{x}) \| p_\theta(\mathbf{z})) - \mathbb{E}_{\mathbf{z}\sim q_\phi(\mathbf{z}\vert\mathbf{x})}\log p_\theta(\mathbf{x}\vert\mathbf{z}) \notag
\end{align}

Then rearrange the left and right-hand side of the equation. We have Eq (\ref{eq:kl_dvergence_2}); moreover, then the loss function would be as the variational lower bound, or evidence lower bound, as shown in Eq (\ref{eq:kl_dvergence_3}).

\begin{align}
\label{eq:kl_dvergence_2}
& \log p_\theta(\mathbf{x}) - D_\text{KL}( q_\phi(\mathbf{z}\vert\mathbf{x}) \| p_\theta(\mathbf{z}\vert\mathbf{x}) ) \\
&= \mathbb{E}_{\mathbf{z}\sim q_\phi(\mathbf{z}\vert\mathbf{x})}\log p_\theta(\mathbf{x}\vert\mathbf{z}) - D_\text{KL}(q_\phi(\mathbf{z}\vert\mathbf{x}) \| p_\theta(\mathbf{z}))  \notag
\end{align}

\begin{align}
\label{eq:kl_dvergence_3}
L_\text{VAE}(\theta, \phi) &= -\log p_\theta(\mathbf{x}) + D_\text{KL}( q_\phi(\mathbf{z}\vert\mathbf{x}) \| p_\theta(\mathbf{z}\vert\mathbf{x}) ) \\
&= - \mathbb{E}_{\mathbf{z} \sim q_\phi(\mathbf{z}\vert\mathbf{x})} p_\theta(\mathbf{x}\vert\mathbf{z}) + D_\text{KL}( q_\phi(\mathbf{z}\vert\mathbf{x}) \| p_\theta(\mathbf{z}) ) \notag\\
\theta^{*}, \phi^{*} &= \arg\min_{\theta, \phi} L_\text{VAE} \notag
\end{align}

Therefore by minimizing the loss, we are maximizing the lower bound of the probability of generating real data samples in Eq (\ref{eq:kl_dvergence_4}).

\begin{align}
\label{eq:kl_dvergence_4}
-L_\text{VAE} = \log p_\theta(\mathbf{x}) - D_\text{KL}( q_\phi(\mathbf{z}\vert\mathbf{x}) \| p_\theta(\mathbf{z}\vert\mathbf{x}) ) \leq \log p_\theta(\mathbf{x})
\end{align}

\section{Proposed Method}
\begin{figure}[H]
\centering
\centerline{\includegraphics[width=1.0\textwidth]{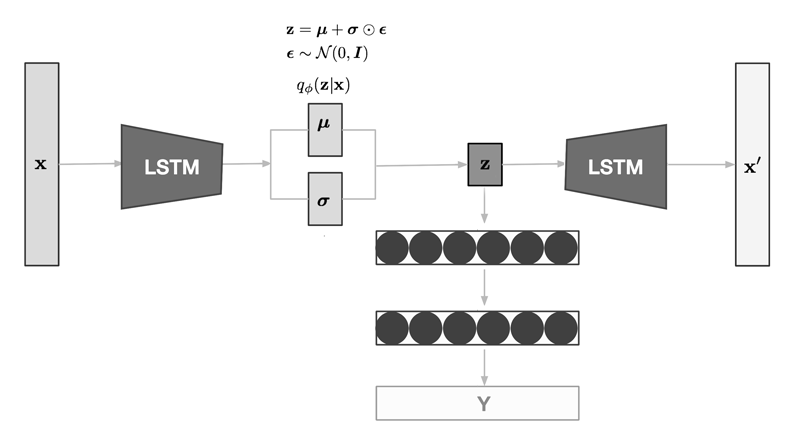}}
\caption{Illustration of the proposed model architecture.}
\label{fig:proposed_model}
\end{figure}

According to the previous approaches, the proposed model includes a Variational Autoencoder, which uses LSTM as its encoder and decoder parts, as shown in Fig (\ref{fig:proposed_model}). Long Short-Term Memory acts as an exploiter both the past and future information — finally, a multi-layer perceptron (MLP) network, which is responsible for mapping the target with the samples of distribution, which learned by the VLSTM-E.

In this proposed approach, the network simultaneously learns the distribution of $z$ and transmits samplings from the distribution and feed into the Multilayer Perceptron model to estimate traffic flow

\section{Experiments}
\subsection{Dataset}

\begin{figure}[H]
\centering
\includegraphics[width=1.0\textwidth]{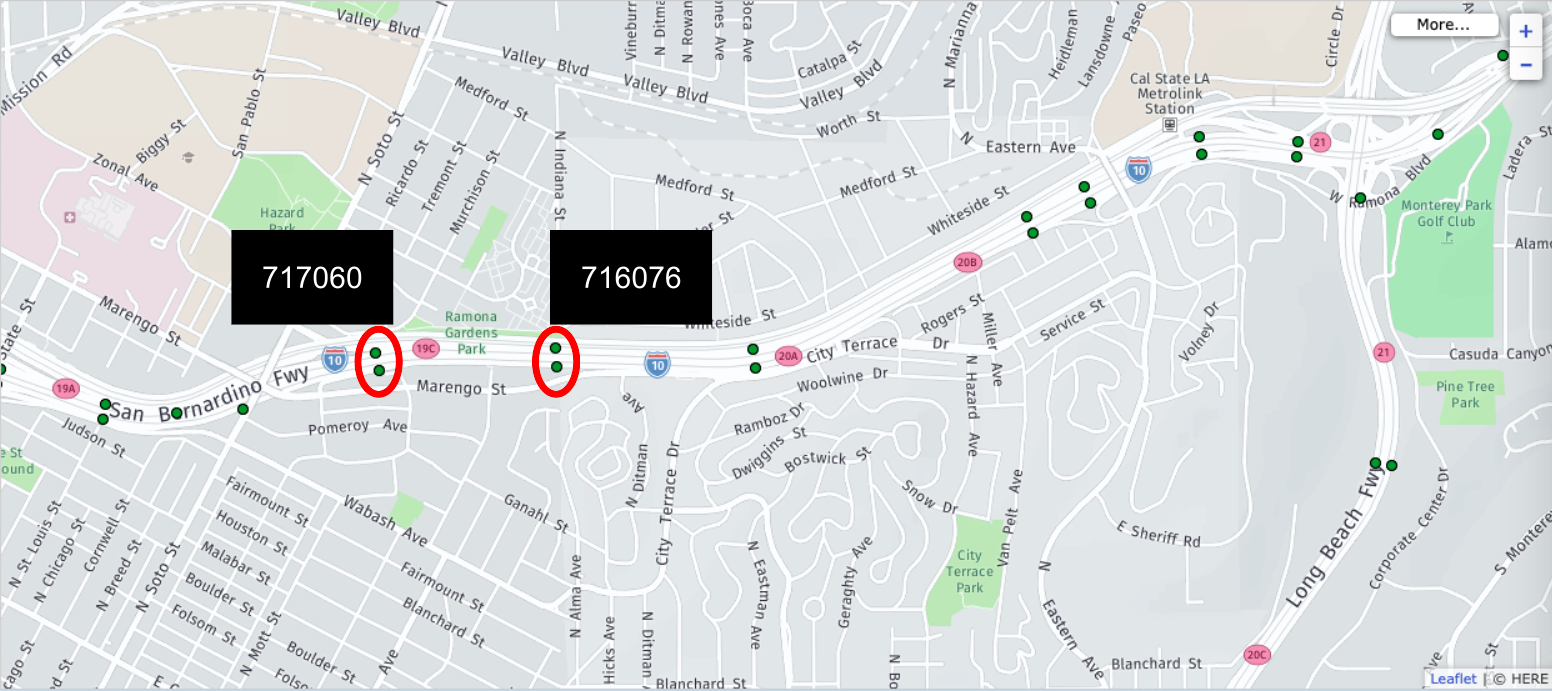}
\caption{The traffic flow between two station in the San Bernardino Fwy.}
\label{fig:dataset_locations}
\end{figure}

Caltrans Performance Measurement System (PeMS) used as a public dataset. It was collected in the real-time form of data by more than 39,000 individual detectors across all major metropolitan areas of the state of California. Performance Measurement System provides a significant variety source of traffic data integrated from Caltrans and other local agency systems.

In this paper, the traffic flow dataset consists of sensors information in the California area, district seven, between 2019-01-01 to 2019-05-30 in a five minutes interval detections. In the case of sensors failure, some records have no values (missing data). 
In this scenario, a combination of Spline-Interpolation and average over a 15 minutes interval, could help the model learn inner patterns desirably. Then the dataset prepared in preprocessing steps. In this particular case, the proposed model would be tested on the traffic flows of two points between station 716076 and 717060, as shown in Fig (\ref{fig:dataset_locations}).

\begin{figure}[t!]
     \centering
     \begin{subfigure}[!h]{0.8\textwidth}
         \centering
         \includegraphics[width=1.0\textwidth]{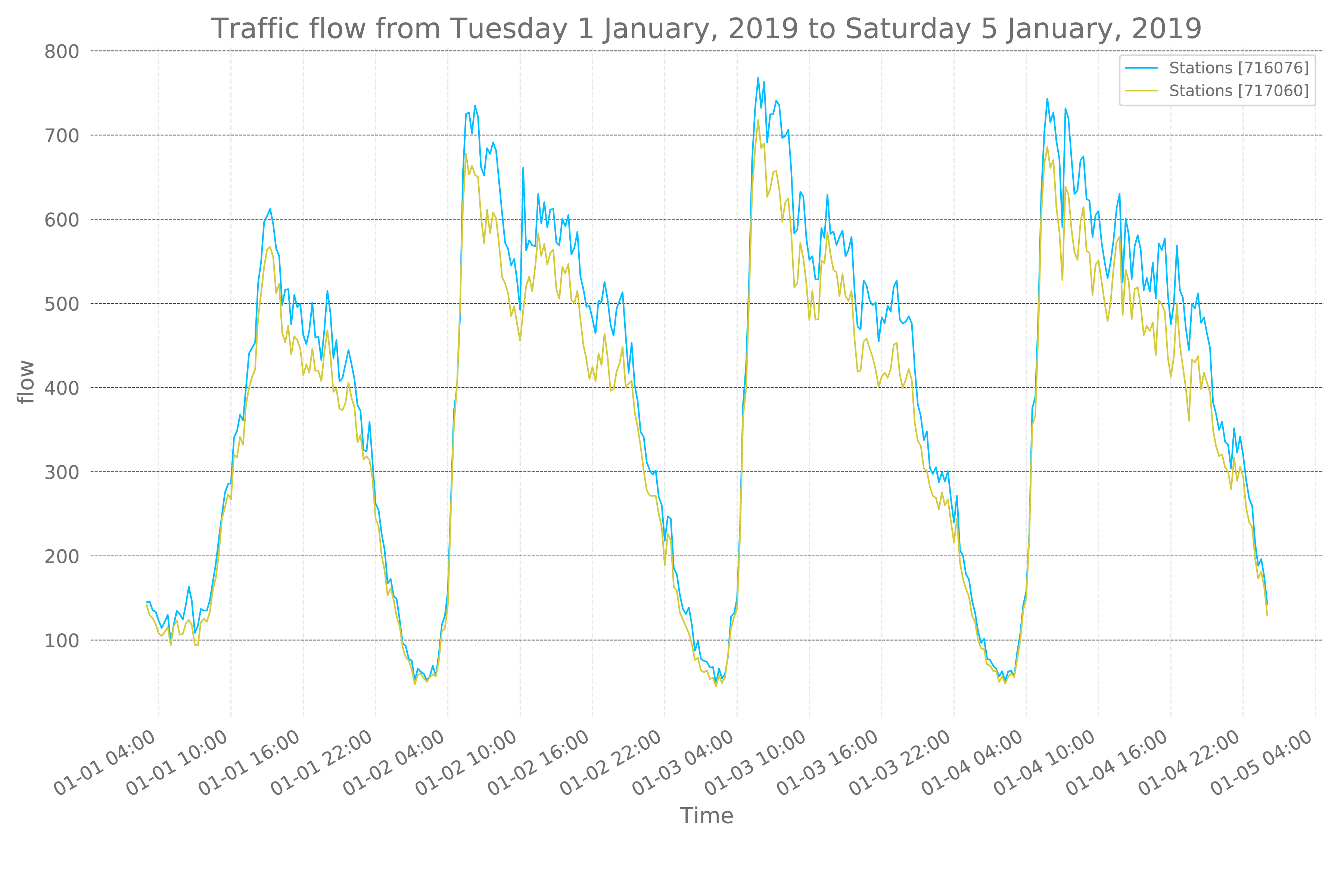}
         \caption{}
     \end{subfigure}
     \hfill
     \begin{subfigure}[!h]{0.8\textwidth}
         \centering
         \includegraphics[width=1.0\textwidth]{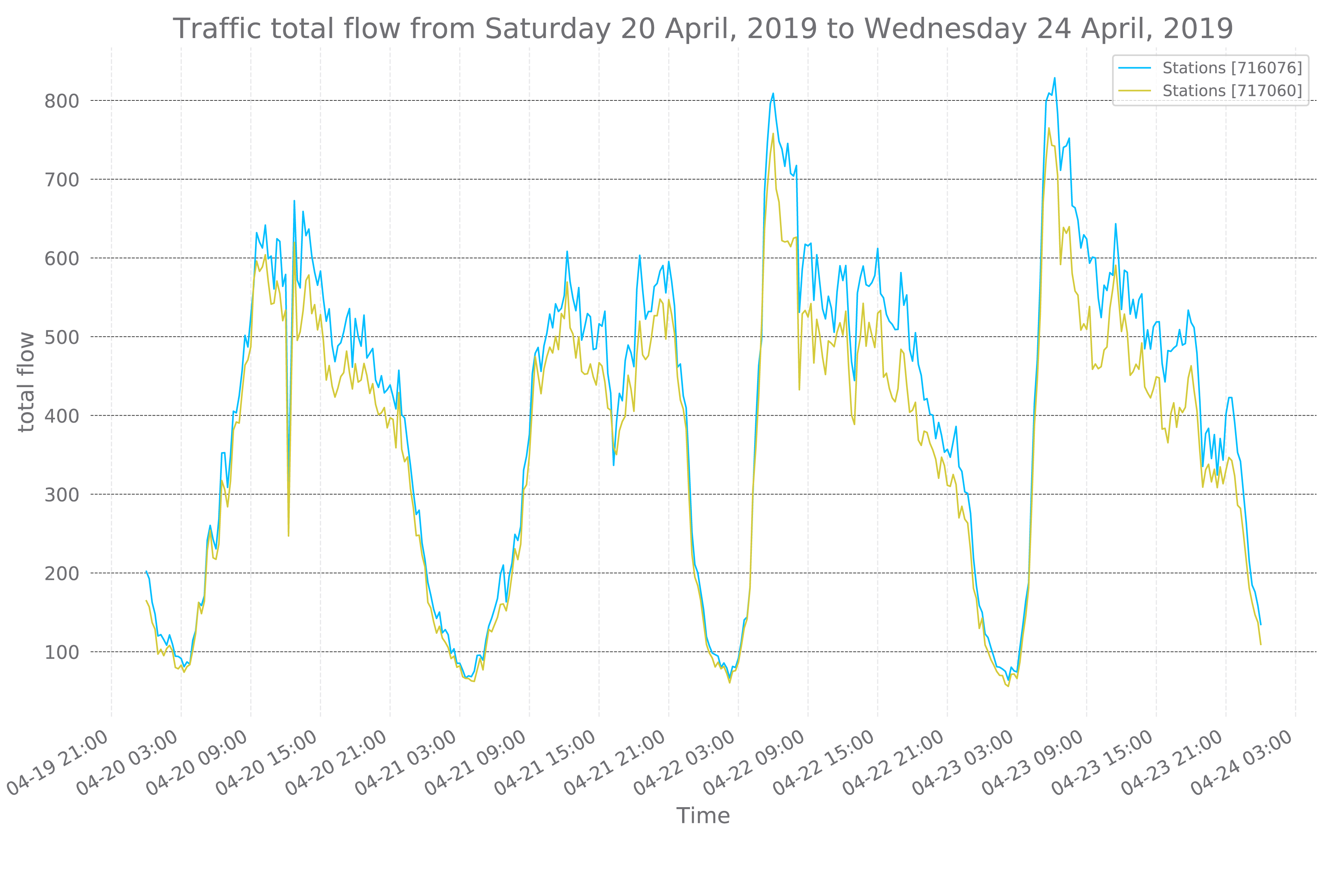}
         \caption{}
     \end{subfigure}
    \caption{Typical daily traffic flow pattern for two stations 716076 and 717060. (a) Traffic flow from Tuesday 1 January 2019 to Saturday 5 January 2019 as a training example. (b) Traffic flow from Saturday 20 April 2019 to Wednesday 24 April 2019 as a testing example.}
    \label{fig:dataset_plot}
\end{figure}

Then for each record at time $t$, data related to time $t_{12}$ is selected as additional features. In other words, our data is picked up to 12 earlier records as a look back. Then the data is scaled into a Min-Max scaler. The data in 2019 between 2019-01-01 00:00:00 to 2019-03-31 23:59:00 chose as a training set others for testing, as shown in Table (\ref{table:train_test_split}). Besides, typical daily traffic flow charts are presented in Fig (\ref{fig:dataset_plot}) for both training and testing parts regarding two stations.

\begin{table}[!htbp]
    \captionsetup{font=scriptsize}
    \caption{Displays the dimensional division of data into training and testing}
    \label{table:train_test_split}
    \footnotesize
    \centering
    \begin{tabular}{lccccl}
        \toprule
        Stations & X Train & Y Train & X Test & Y Test  \\
        \midrule
        716076  & 8628 x 12 x 1 & 5778 x 12 x 1 & 8628 x 1 & 5778 x 1  \\
        717060  & 8628 x 12 x 1 & 6187 x 12 x 1 & 8628 x 1 & 6187 x 1\\
        \bottomrule
    \end{tabular}
\end{table}

\subsection{Parametric Settings}

In terms of hardware, the GPU we use is Tesla k80 which provided by Google Colab\cite{google}. The proposed VLSTM-E architecture and chosen networks were implemented on the TensorFlow platform (v1.14.0)  \cite{tensorflow}. The learning rate is 0.0001, and the batch size is 256, the sigmoid is used for both as the activation of the last layer.

\subsection{Index of Performance}
Four measurements introduced in this paper to evaluate the effectiveness of the proposed model, in the follows:

\begin{equation}
e_{i} =  f_{i} -  \widehat{f_{i}}\label{e}
\end{equation}

\begin{equation}
MSE = \frac{1}{n}\sum_{t=1}^{n}e_{i}^2\label{mse}
\end{equation}

\begin{equation}
RMSE = \displaystyle\sqrt{\frac{1}{n}\sum_{t=1}^{n}e_{i}^2}\label{rmse}
\end{equation}

\begin{equation}
MAE = \dfrac{1}{n}\sum\limits_{t=1}^{n}|e_{i}|\label{mae}
\end{equation}

\begin{equation}
MAPE = \displaystyle\frac{100\%}{n}\sum_{t=1}^{n}\left |\frac{e_{i}}{f_i}\right|\label{mape}
\end{equation}

where n is the number of the test sample, $f_{i}$ is the real traffic flow in sample $i$, and $\widehat{f_{i}}$ denotes the predicted traffic flow. 

\section{Results}
\begin{figure*}[!htp]
     \centering
     \begin{subfigure}[!h]{0.8\textwidth}
         \centering
         \includegraphics[width=1.0\textwidth]{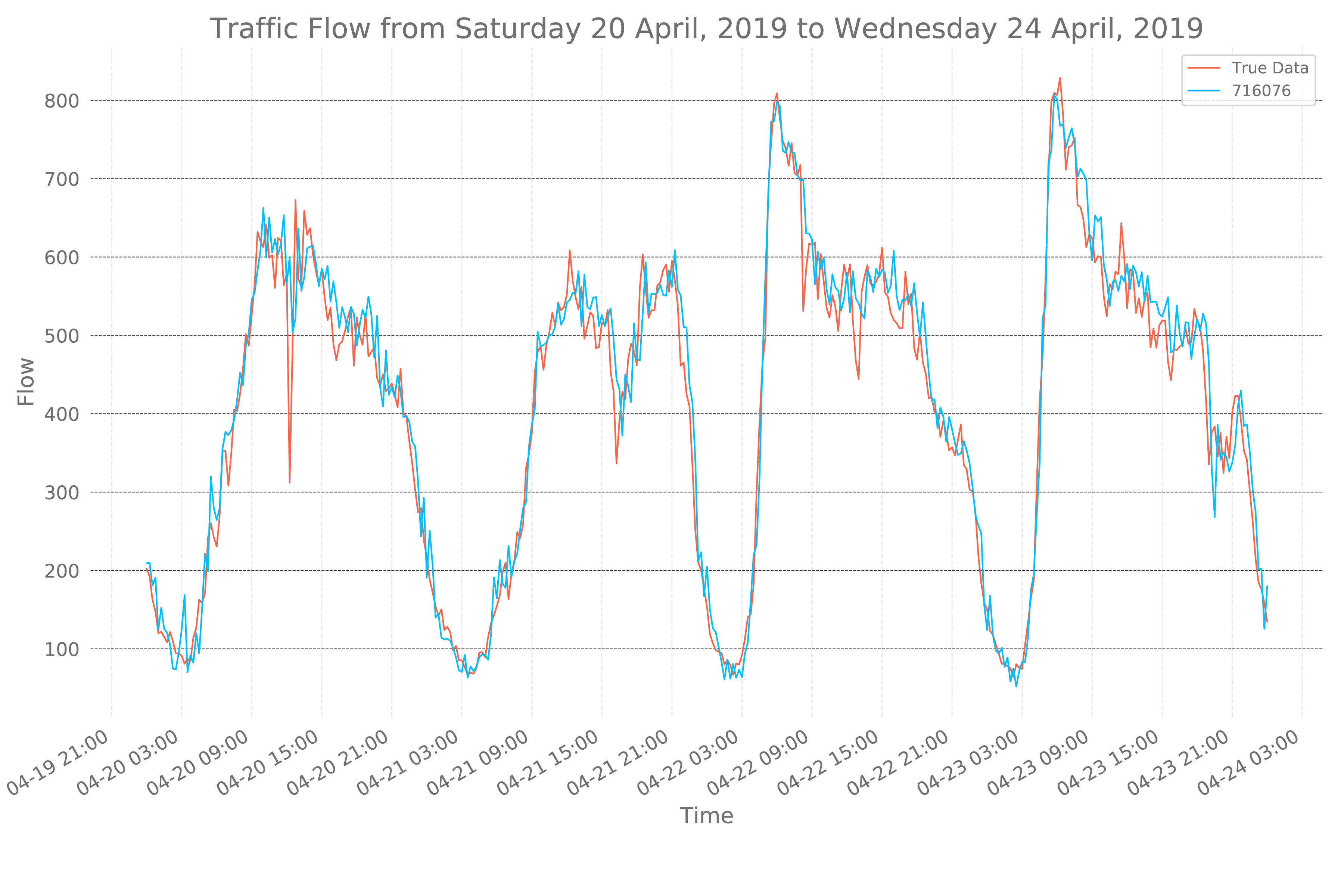}
         \caption{}
     \end{subfigure}
     \hfill
     \begin{subfigure}[!h]{0.8\textwidth}
         \centering
         \includegraphics[width=1.0\textwidth]{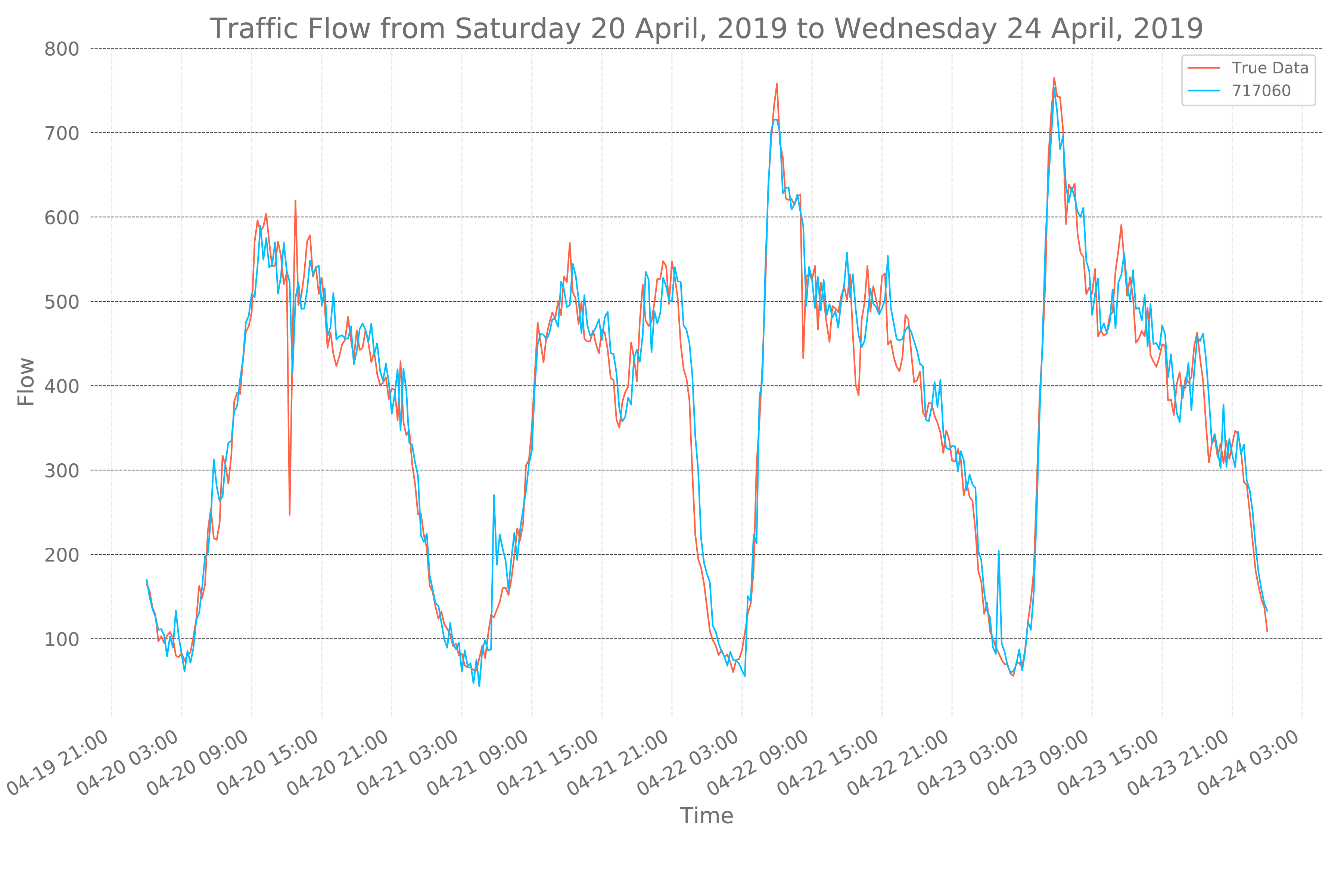}
         \caption{}
     \end{subfigure}
    \caption{Typical daily traffic flow forecasting for two stations 716076 and 717060 by VLSTM-E model between Saturday 20 April 2019 to Wednesday 24 April 2019. (a) Traffic flow forecasting for 716076. (b) Traffic flow forecasting for 717060.}
    \label{fig:vle_results_plot}
\end{figure*}

\begin{figure*}[!htp]
     \centering
     \begin{subfigure}[!h]{0.8\textwidth}
         \centering
         \includegraphics[width=1.0\textwidth]{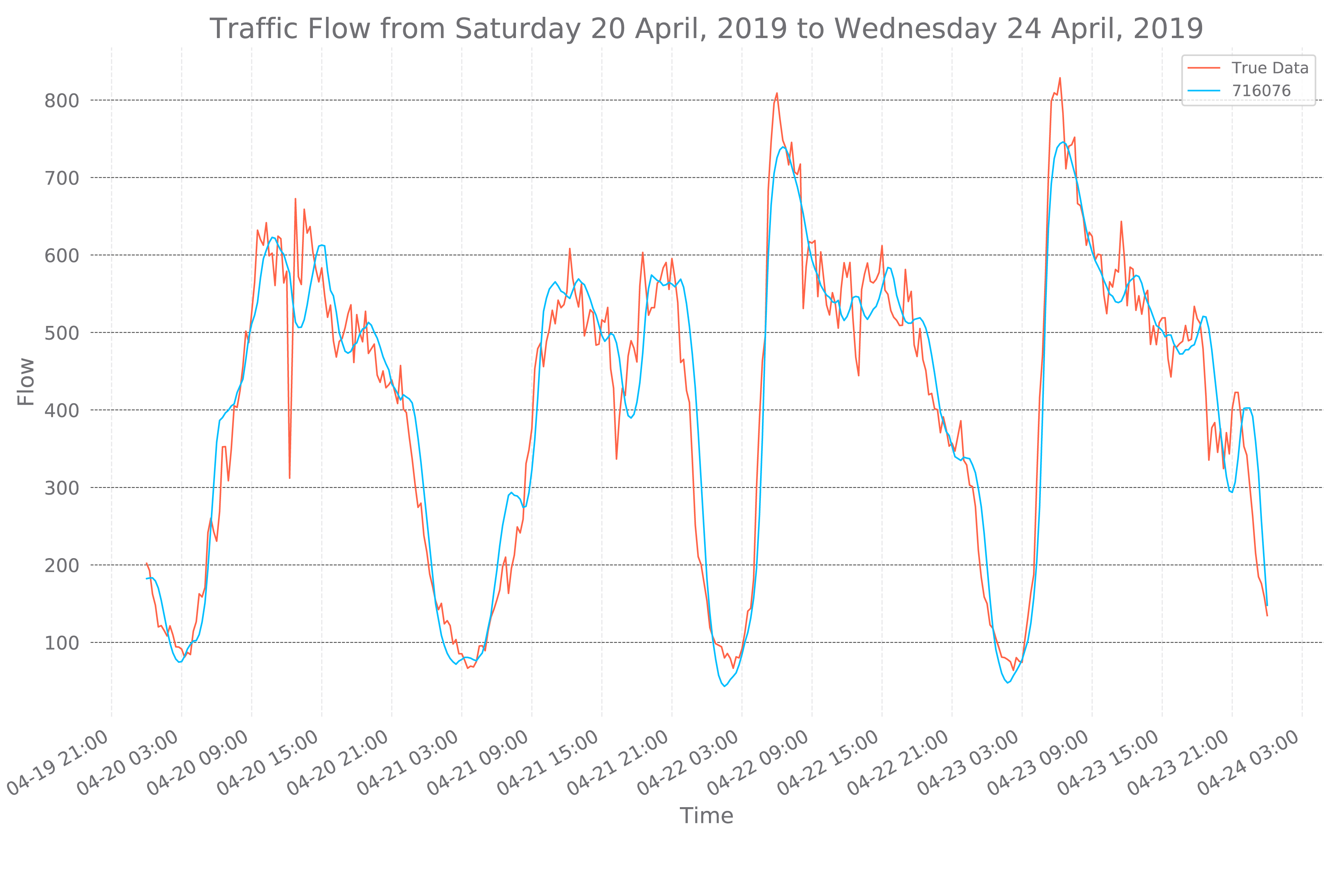}
         \caption{}
     \end{subfigure}
     \hfill
     \begin{subfigure}[!h]{0.8\textwidth}
         \centering
         \includegraphics[width=1.0\textwidth]{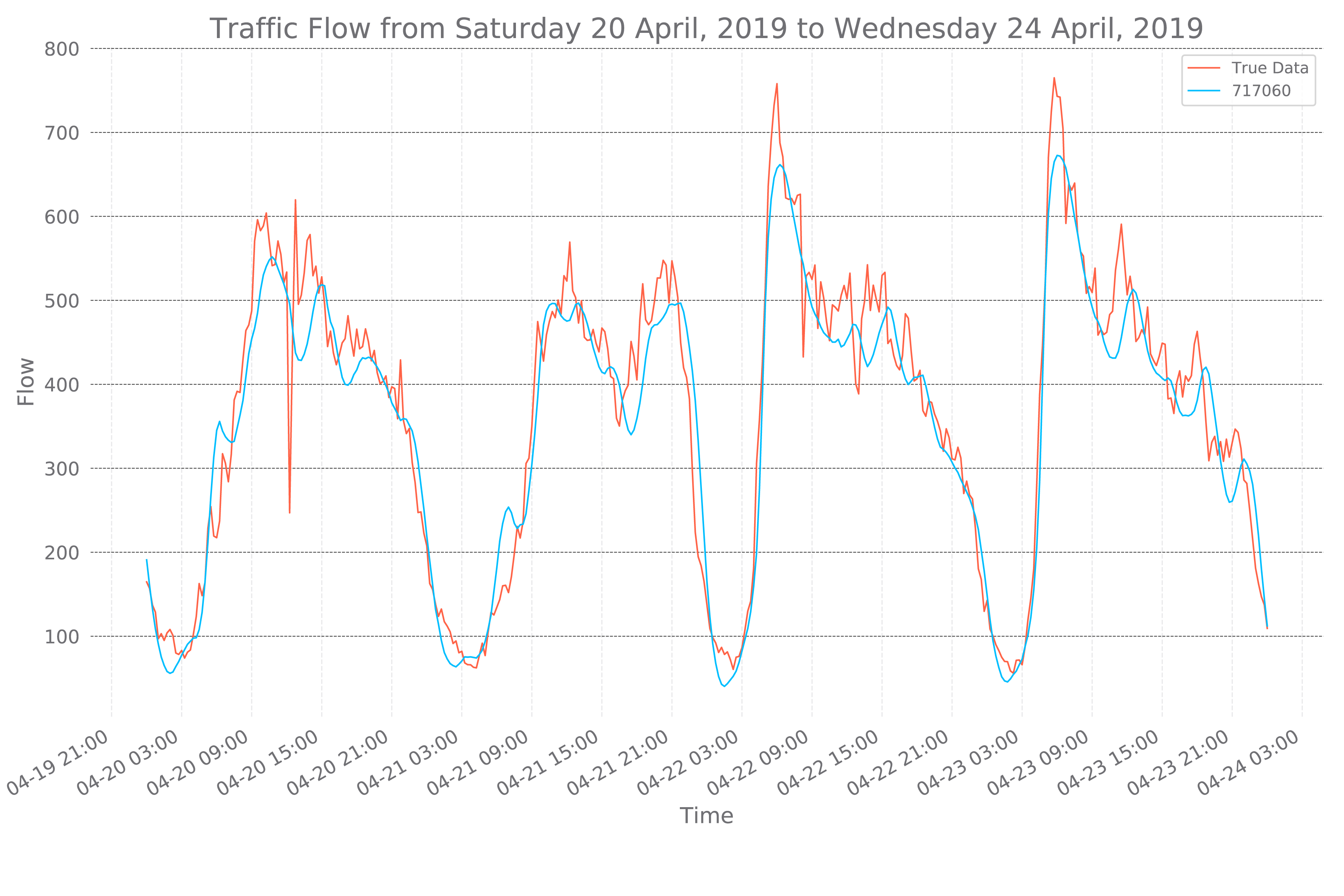}
         \caption{}
     \end{subfigure}
    \caption{Typical daily traffic flow forecasting for two stations 716076 and 717060 by LSTM model between Saturday 20 April 2019 to Wednesday 24 April 2019. (a) Traffic flow forecasting for 716076. (b) Traffic flow forecasting for 717060.}
    \label{fig:lstm_results_plot}
\end{figure*}

\begin{figure*}[!htbp]
     \centering
     \begin{subfigure}[!h]{0.8\textwidth}
         \centering
         \includegraphics[width=1.0\textwidth]{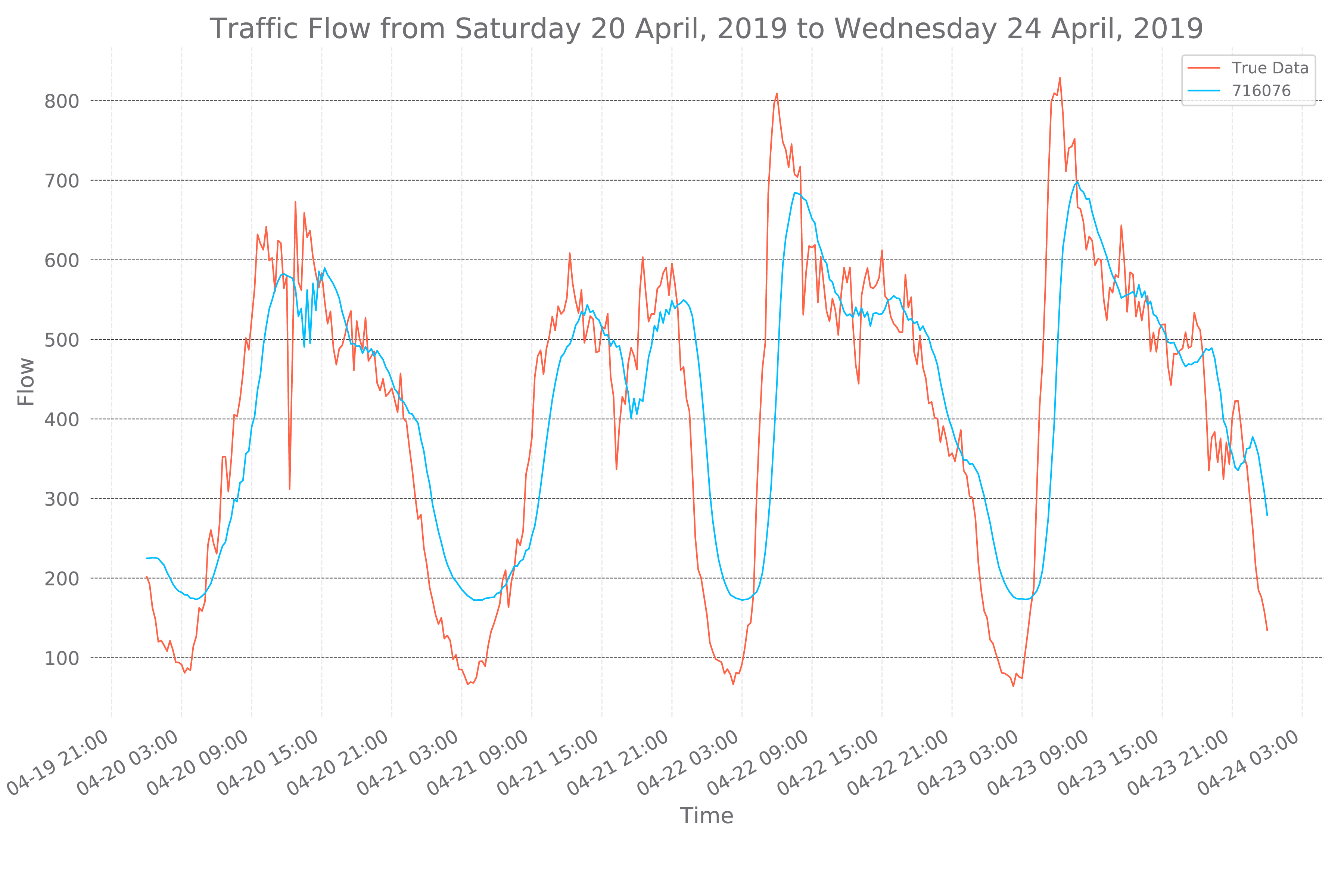}
         \caption{}
     \end{subfigure}
     \hfill
     \begin{subfigure}[!h]{0.8\textwidth}
         \centering
         \includegraphics[width=1.0\textwidth]{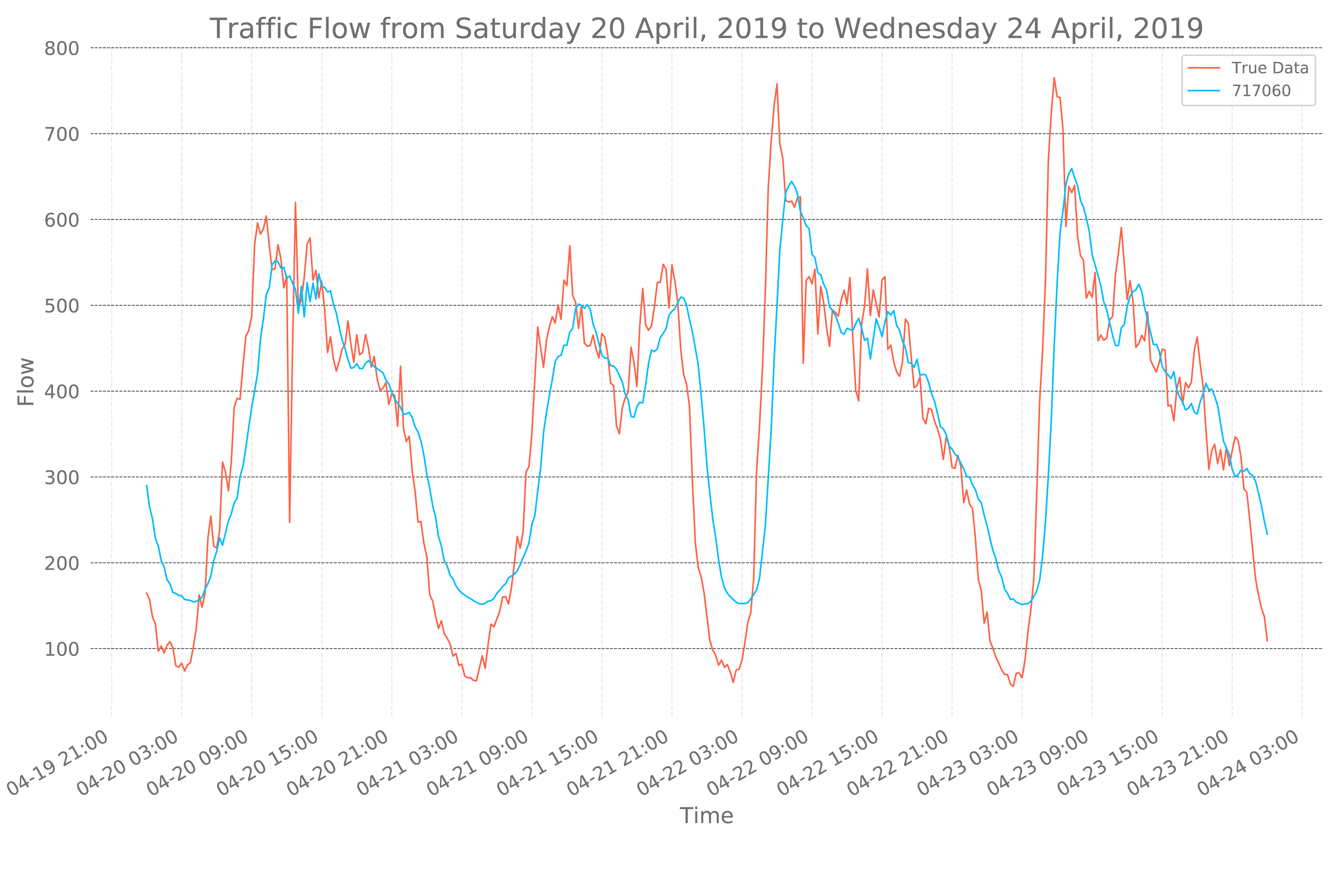}
         \caption{}
     \end{subfigure}
    \caption{Typical daily traffic flow forecasting for two stations 716076 and 717060 by MCNNM model between Saturday 20 April 2019 to Wednesday 24 April 2019. (a) Traffic flow forecasting for 716076. (b) Traffic flow forecasting for 717060.}
    \label{fig:mcnnm_results_plot}
\end{figure*}

\begin{figure*}[!ht]
     \centering
     \begin{subfigure}[!h]{0.8\textwidth}
         \centering
         \includegraphics[width=1.0\textwidth]{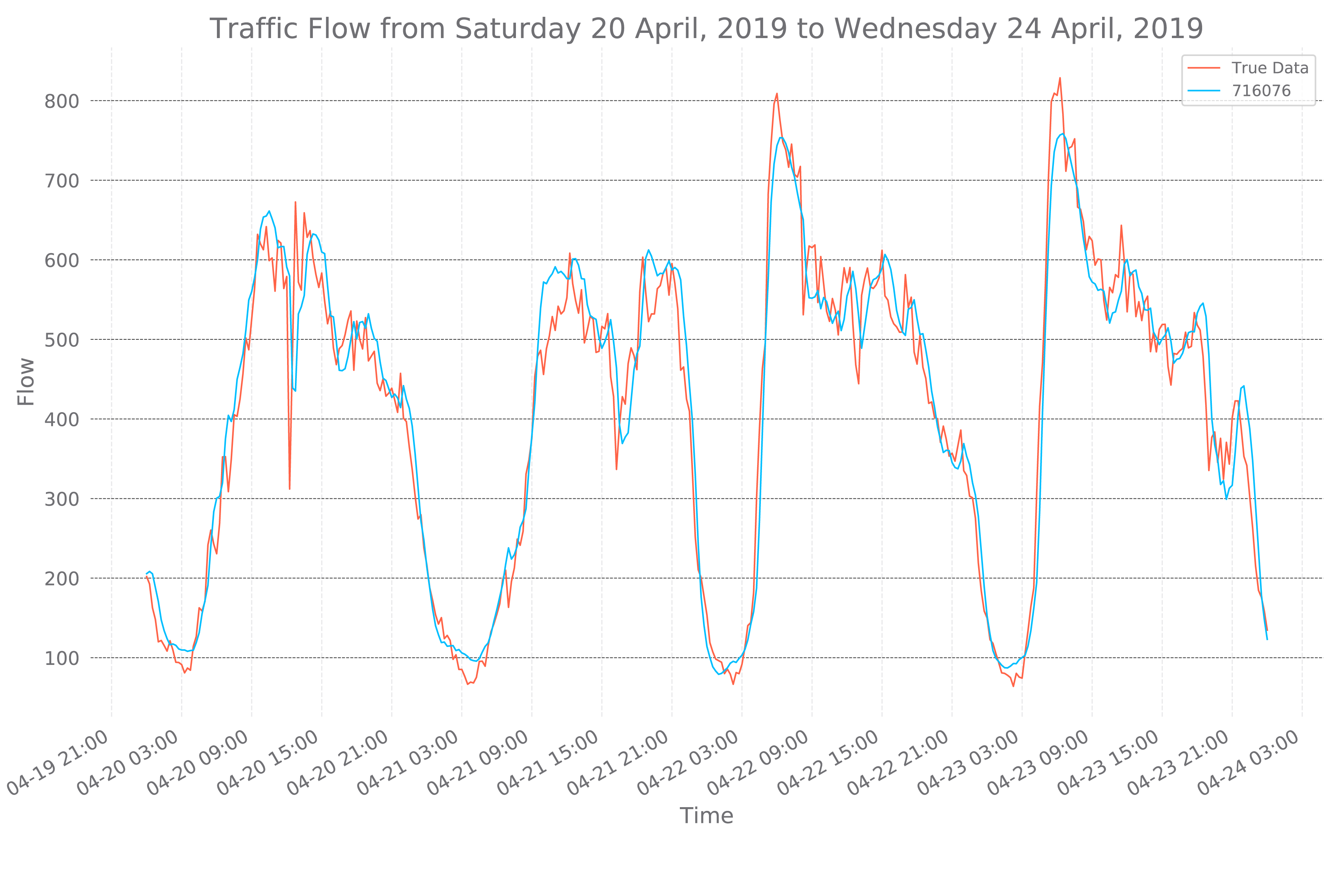}
         \caption{}
     \end{subfigure}
     \hfill
     \begin{subfigure}[!h]{0.8\textwidth}
         \centering
         \includegraphics[width=1.0\textwidth]{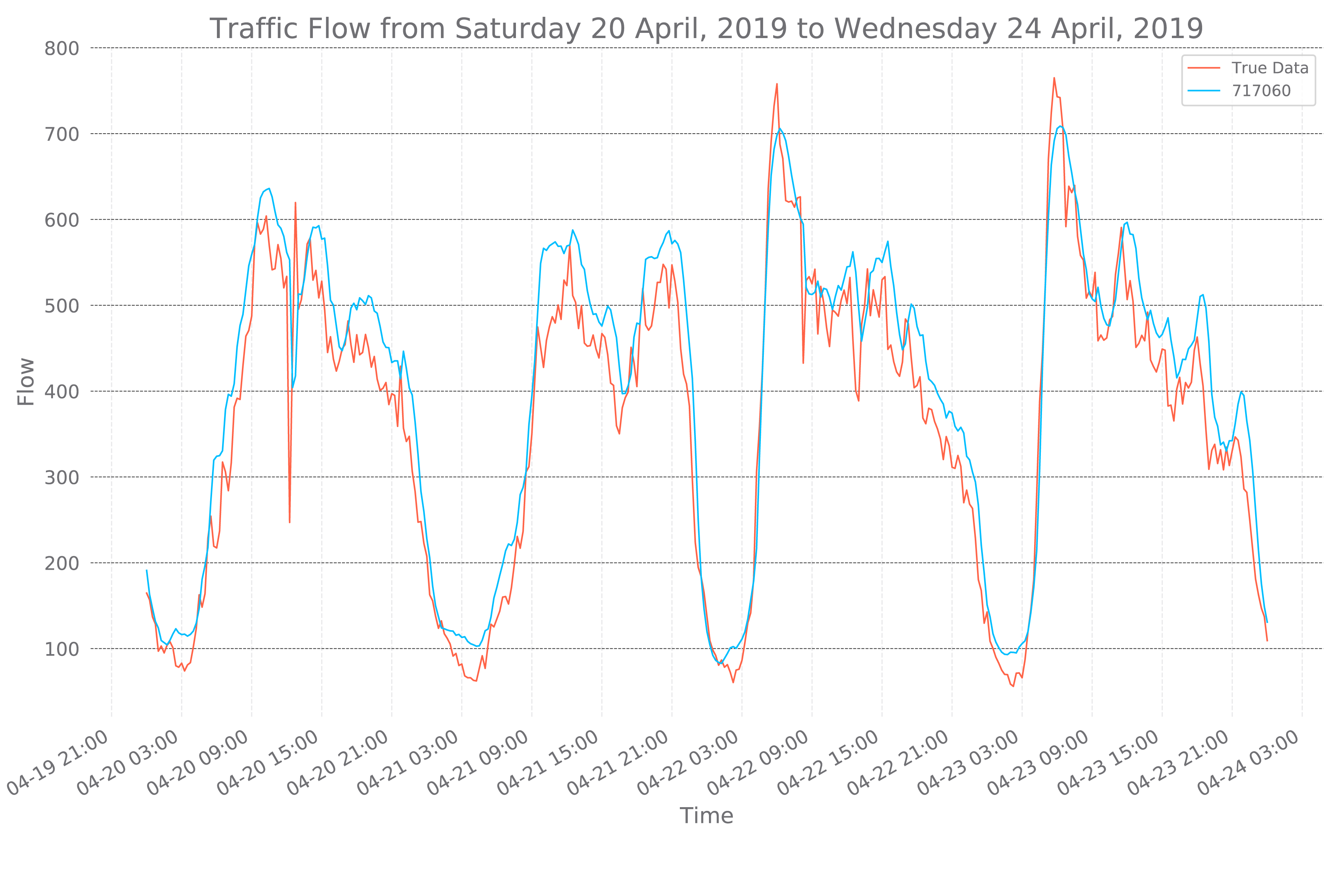}
         \caption{}
     \end{subfigure}
    \caption{Typical daily traffic flow forecasting for two stations 716076 and 717060 by SAEs model between Saturday 20 April 2019 to Wednesday 24 April 2019. (a) Traffic flow forecasting for 716076. (b) Traffic flow forecasting for 717060.}
    \label{fig:saes_results_plot}
\end{figure*}

In the following, the results presented as evaluation results and forecasting the traffic flow for VLSTM-E (Table (\ref{table:vle_results}), Fig (\ref{fig:vle_results_plot})), LSTM (Table (\ref{table:lstm_results}), Fig (\ref{fig:lstm_results_plot})), MCNNM (Table (\ref{table:mcnnm_results}), Fig (\ref{fig:mcnnm_results_plot})), and SAEs (Table (\ref{table:saes_results}), Fig (\ref{fig:saes_results_plot})), respectively.

\begin{threeparttable}[H]
    \small
    \caption{The evaluation results for the Variational Long Short-Term Memory Encoder (VLSTM-E) model.}
    \label{table:vle_results}
    \setlength\tabcolsep{0pt}
    \begin{tabular*}{\linewidth}{@{\extracolsep{\fill}} l cc cc cc @{}}
        \toprule &   \multicolumn{5}{c}{VLSTM-E} \\ \cmidrule{2-5}
        Station ID          & MAPE [\%] & MAE & MSE & RMSE \\
        \midrule
    716076          & 9.5954 & 0.0312 & 0.0018 & 0.0422 \\
    717060          & 8.8625 & 0.0276 & 0.0015 & 0.0381 \\
    \bottomrule
    \end{tabular*}
    \vspace{5mm}
\end{threeparttable}

\begin{threeparttable}[H]
    \small
    \caption{The evaluation results for the Long Short-Term Memory (LSTM) model.}
    \label{table:lstm_results}
    \setlength\tabcolsep{0pt}
    \begin{tabular*}{\linewidth}{@{\extracolsep{\fill}} l cc cc cc @{}}
        \toprule &   \multicolumn{5}{c}{LSTM} \\ \cmidrule{2-5}
        Station ID          & MAPE [\%] & MAE & MSE & RMSE \\
        \midrule
    716076          & 10.2718 & 0.0341 & 0.0024 & 0.0490 \\
    717060          & 10.8174 & 0.0366 & 0.0022 & 0.0464 \\
    \bottomrule
    \end{tabular*}
    \vspace{5mm}
\end{threeparttable}

\begin{threeparttable}[H]
    \small
    \caption{The evaluation results for the Multiple Convolutional Neural Network for Multivariate (MCNNM) model.}
    \label{table:mcnnm_results}
    \setlength\tabcolsep{0pt}
    \begin{tabular*}{\linewidth}{@{\extracolsep{\fill}} l cc cc cc @{}}
        \toprule &   \multicolumn{5}{c}{MCNNM} \\ \cmidrule{2-5}
        Station ID          & MAPE [\%] & MAE & MSE & RMSE \\
        \midrule
    716076          & 31.0840 & 0.0757 & 0.0129 & 0.1136 \\
    717060          & 24.0724 & 0.0603 & 0.0082 & 0.0905 \\
    \bottomrule
    \end{tabular*}
    \vspace{5mm}
\end{threeparttable}

\begin{threeparttable}[H]
    \small
    \caption{The evaluation results for the Stacked Autoencoders (SAEs) model.}
    \label{table:saes_results}
    \setlength\tabcolsep{0pt}
    \begin{tabular*}{\linewidth}{@{\extracolsep{\fill}} l cc cc cc @{}}
        \toprule &   \multicolumn{5}{c}{SAEs} \\ \cmidrule{2-5}
        Station ID          & MAPE [\%] & MAE & MSE & RMSE \\
        \midrule
    716076          & 9.9421 & 0.0326 & 0.0020 & 0.0449 \\
    717060          & 18.4939 & 0.0560 & 0.0040 & 0.0635 \\
    \bottomrule
    \end{tabular*}
    \vspace{5mm}
\end{threeparttable}

As the results show, the proposed model, VLSTM-E, has improved compared to other conventional models like the Stacked Autoencoders, Long Short-Term Memory, and Multiple Convolutional Neural Network, which introduced in 2015 \cite{Lv2014}, 2016 \cite{Fu2016} and 2019 \cite{Wang2019}. To better understanding, this superiority, the average of the results according to the evaluation criterion is presented in Table (\ref{table:model_avg}) which, shows the MSE score of the VLSTM-E is 0.0016. 

\begin{threeparttable}[H]
\small
\caption{Average performance for all the models.}
\label{table:model_avg}
\setlength\tabcolsep{0pt}
\begin{tabular*}{\linewidth}{@{\extracolsep{\fill}} l cc cc cc @{}}
\toprule
&   \multicolumn{5}{c}{Average Models} \\ \cmidrule{2-5}
Station ID          & MAPE [\%] & MAE & MSE & RMSE \\
\midrule
VLSTM-E                     & 9.2290 & 0.0294 & 0.0016 & 0.0402 \\
LSTM \cite{Fu2016}          & 10.5446 & 0.0353 & 0.0023 & 0.0477 \\
MCNNM \cite{Wang2019}       & 27.5782 & 0.0680 & 0.0106 & 0.1021 \\
SAEs \cite{Lv2014}          & 14.2180 & 0.0443 & 0.0030 & 0.0542 \\
\bottomrule
\end{tabular*}
\vspace{5mm}
\end{threeparttable}

\begin{figure*}[!htp]
\centering
\includegraphics[width=1.0\textwidth]{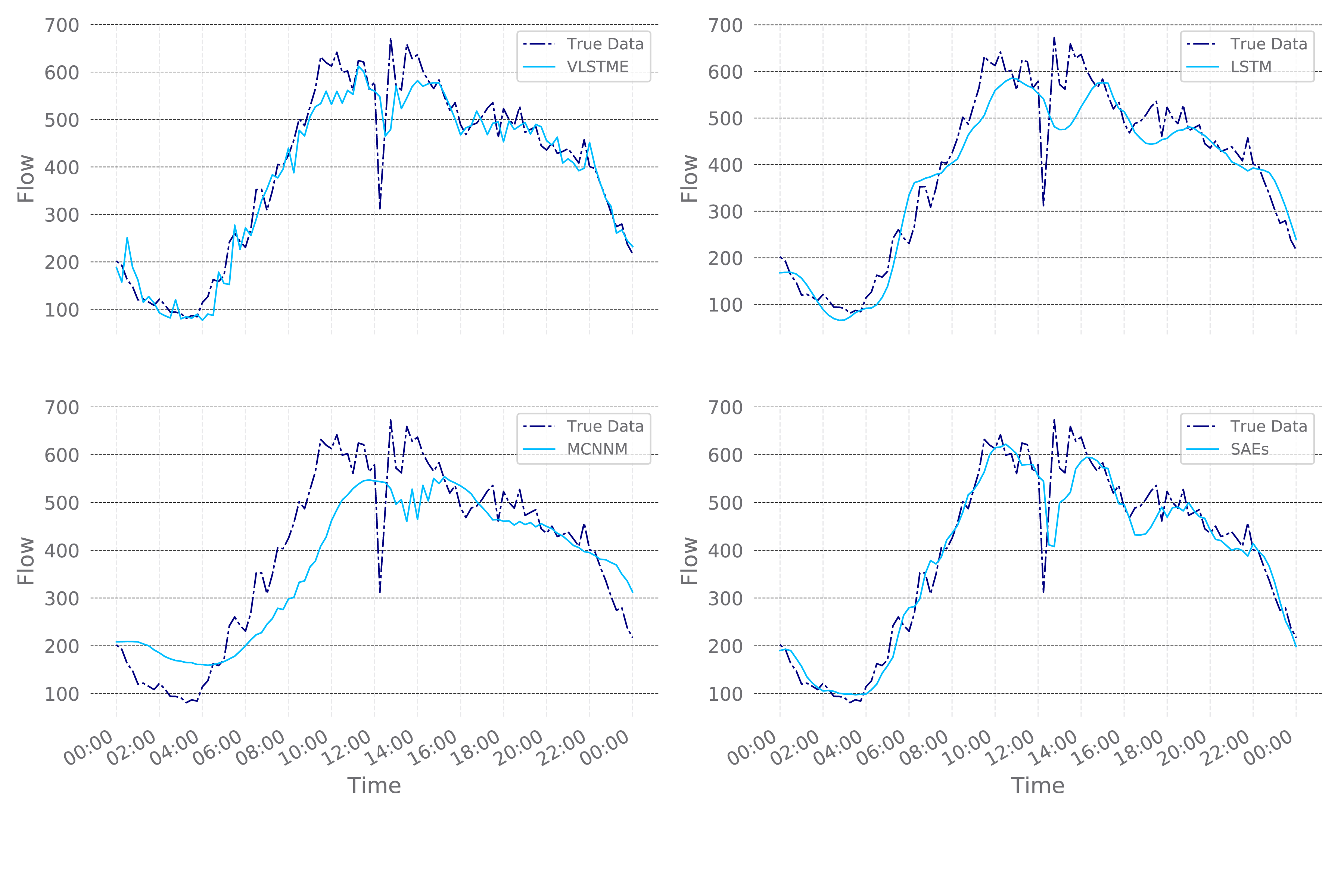}
\caption{Forecasting performance on Varitional Long Short-Term Memory Encoder (VLSTME), Long Short-Term Memory (LSTM), Multiple Convolutional Neural Network for Multivariate (MCNNM), and Stacked Autoencoders (SAEs) for 716076 station!}
\label{fig:flow_prediction_1}
\end{figure*}

\begin{figure*}[!htp]
\centering
\includegraphics[width=1.0\textwidth]{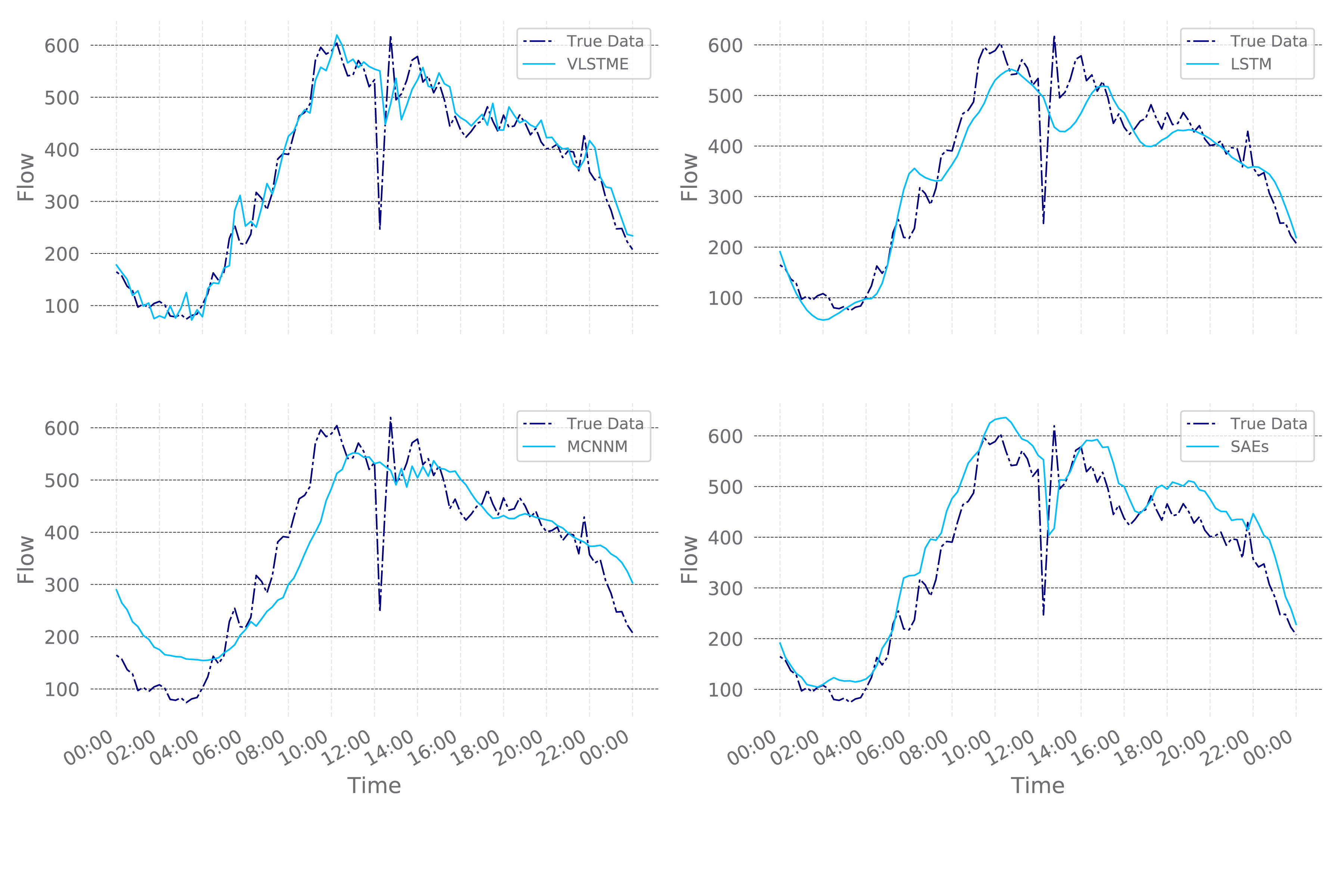}
\caption{Forecasting performance on Varitional Long Short-Term Memory Encoder (VLSTME), Long Short-Term Memory (LSTM), Multiple Convolutional Neural Network for Multivariate (MCNNM), and Stacked Autoencoders (SAEs) for 717060 station!}
\label{fig:flow_prediction_2}
\end{figure*}

Figures (\ref{fig:flow_prediction_1}, \ref{fig:flow_prediction_2}) shows the prediction results for the two stations 716076, and 717060 for the test dataset on 2019, April 20. As can be seen, in all stations, the VLSTM-E curve has a better estimation of the traffic flow than other curves. In cases where the traffic flow fluctuates in viewing a large amount of traffic, the model can quickly converge into that behavior. Also, in low volume volatility, imitation shows a better response than the Long Short-Term Memory model. Perhaps the reason for this improvement can be found in the data structure; in some cases, the sensors in the stations can not detect the observation, or even this observation will not be highly accurate. In another word, these sensors might be failed in vehicle detection, so it caused missing values. Since the model related to the distribution of data, and the sample of this distribution feed into the network, it can be reduced the adverse effects of these missing data in the learning process and lead to satisfactory results than the other models like Long Short-Term Memory.

\section{Conclusions}
This paper presents a Deep Learning approach with a Variational Long Short-Term Memory Encoder to predict the short-term traffic flow. In contrast to the previous approaches \cite{Fu2016}, this model considers the pattern of the data and provided a solution for missing data. So, it could achieve better results based on the four evaluation criteria in contrast to the other models \cite{Lv2014,Fu2016,Wang2019}, which were introduced earlier. This model is implemented on the PeMS dataset. A suggestion for future work would be interesting if implemented on the other dataset that the stations and its sensors produce missing or low-value information. Also, on various distributions, such as Dirichlet distribution, can be useful in improving sample distribution in traffic flow.

\clearpage
\bibliographystyle{unsrt}  

\begin{thebibliography}{1}

\bibitem{Hou2016RepeatabilityAS}
Zhongsheng Hou and Xingyi Li.
\newblock Repeatability and similarity of freeway traffic flow and long-term
  prediction under big data.
\newblock {\em IEEE Transactions on Intelligent Transportation Systems},
  17:1786--1796, 2016.

\bibitem{Oh2015UrbanTF}
Se~do~Oh, Young jin Kim, and Ji~sun Hong.
\newblock Urban traffic flow prediction system using a multifactor pattern
  recognition model.
\newblock {\em IEEE Transactions on Intelligent Transportation Systems},
  16:2744--2755, 2015.

\bibitem{Stathopoulos2003}
Anthony Stathopoulos and Matthew~G. Karlaftis.
\newblock A multivariate state space approach for urban traffic flow modeling
  and prediction.
\newblock {\em Transportation Research Part C: Emerging Technologies},
  11(2):121--135, April 2003.

\bibitem{Zhou2019}
Teng Zhou, Dazhi Jiang, Zhizhe Lin, Guoqiang Han, Xuemiao Xu, and Jing Qin.
\newblock Hybrid dual kalman filtering model for short-term traffic flow
  forecasting.
\newblock {\em {IET} Intelligent Transport Systems}, 13(6):1023--1032, June
  2019.

\bibitem{Zhang2014}
Yanru Zhang, Yunlong Zhang, and Ali Haghani.
\newblock A hybrid short-term traffic flow forecasting method based on spectral
  analysis and statistical volatility model.
\newblock {\em Transportation Research Part C: Emerging Technologies},
  43:65--78, June 2014.

\bibitem{Krblek2019}
Milan Krb{\'{a}}lek, Ji{\v{r}}{\'{\i}} Apeltauer, and Franti{\v{s}}ek
  {\v{S}}eba.
\newblock Traffic flow merging {\textendash} statistical and numerical modeling
  of microstructure.
\newblock {\em Journal of Computational Science}, 32:99--105, March 2019.

\bibitem{Luo2018}
Xianglong Luo, Liyao Niu, and Shengrui Zhang.
\newblock An algorithm for traffic flow prediction based on improved sarima and
  ga.
\newblock {\em KSCE Journal of Civil Engineering}, 22(10):4107--4115, Oct 2018.

\bibitem{Hou2019}
Qinzhong Hou, Junqiang Leng, Guosheng Ma, Weiyi Liu, and Yuxing Cheng.
\newblock An adaptive hybrid model for short-term urban traffic flow
  prediction.
\newblock {\em Physica A: Statistical Mechanics and its Applications},
  527:121065, August 2019.

\bibitem{Ihueze2018}
Chukwutoo~C. Ihueze and Uchendu~O. Onwurah.
\newblock Road traffic accidents prediction modelling: An analysis of anambra
  state, nigeria.
\newblock {\em Accident Analysis {\&} Prevention}, 112:21--29, March 2018.

\bibitem{Zhu2016}
Guangyu Zhu, Kang Song, Peng Zhang, and Li~Wang.
\newblock A traffic flow state transition model for urban road network based on
  hidden markov model.
\newblock {\em Neurocomputing}, 214:567--574, November 2016.

\bibitem{Zhang2017}
Liguo Zhang and Christophe Prieur.
\newblock Stochastic stability of markov jump hyperbolic systems with
  application to traffic flow control.
\newblock {\em Automatica}, 86:29--37, December 2017.

\bibitem{Huang2007}
Darong Huang and Xing rong Bai.
\newblock A wavelet neural network optimal control model for traffic-flow
  prediction in intelligent transport systems.
\newblock In {\em Advanced Intelligent Computing Theories and Applications.
  With Aspects of Artificial Intelligence}, pages 1233--1244. Springer Berlin
  Heidelberg, 2007.

\bibitem{Agarwal2016}
Shaurya Agarwal, Pushkin Kachroo, and Emma Regentova.
\newblock A hybrid model using logistic regression and wavelet transformation
  to detect traffic incidents.
\newblock {\em {IATSS} Research}, 40(1):56--63, July 2016.

\bibitem{Apronti2016}
Dick Apronti, Khaled Ksaibati, Kenneth Gerow, and Jaime~Jo Hepner.
\newblock Estimating traffic volume on wyoming low volume roads using linear
  and logistic regression methods.
\newblock {\em Journal of Traffic and Transportation Engineering (English
  Edition)}, 3(6):493--506, December 2016.

\bibitem{Cai2016}
Pinlong Cai, Yunpeng Wang, Guangquan Lu, Peng Chen, Chuan Ding, and Jianping
  Sun.
\newblock A spatiotemporal correlative k-nearest neighbor model for short-term
  traffic multistep forecasting.
\newblock {\em Transportation Research Part C: Emerging Technologies},
  62:21--34, January 2016.

\bibitem{Sharma2016}
A.~Sharma, R.~Vijay, G.~L. Bodhe, and L.~G. Malik.
\newblock An adaptive neuro-fuzzy interface system model for traffic
  classification and noise prediction.
\newblock {\em Soft Computing}, 22(6):1891--1902, November 2016.

\bibitem{Guo2018}
Jianhua Guo, Zhao Liu, Wei Huang, Yun Wei, and Jinde Cao.
\newblock Short-term traffic flow prediction using fuzzy information
  granulation approach under different time intervals.
\newblock {\em {IET} Intelligent Transport Systems}, 12(2):143--150, March
  2018.

\bibitem{Chen2018}
Weihong Chen, Jiyao An, Renfa Li, Li~Fu, Guoqi Xie, Md~Zakirul~Alam Bhuiyan,
  and Keqin Li.
\newblock A novel fuzzy deep-learning approach to traffic flow prediction with
  uncertain spatial{\textendash}temporal data features.
\newblock {\em Future Generation Computer Systems}, 89:78--88, December 2018.

\bibitem{Goves2016}
Carl Goves, Robin North, Ryan Johnston, and Graham Fletcher.
\newblock Short term traffic prediction on the {UK} motorway network using
  neural networks.
\newblock {\em Transportation Research Procedia}, 13:184--195, 2016.

\bibitem{Raj2016}
Jithin Raj, Hareesh Bahuleyan, and Lelitha~Devi Vanajakshi.
\newblock Application of data mining techniques for traffic density estimation
  and prediction.
\newblock {\em Transportation Research Procedia}, 17:321--330, 2016.

\bibitem{Li2017}
Kui-Lin Li, Chun-Jie Zhai, and Jian-Min Xu.
\newblock Short-term traffic flow prediction using a methodology based on
  {ARIMA} and {RBF}-{ANN}.
\newblock In {\em 2017 Chinese Automation Congress ({CAC})}. {IEEE}, October
  2017.

\bibitem{Sharma2018}
Bharti Sharma, Sachin Kumar, Prayag Tiwari, Pranay Yadav, and Marina~I.
  Nezhurina.
\newblock {ANN} based short-term traffic flow forecasting in undivided two lane
  highway.
\newblock {\em Journal of Big Data}, 5(1), December 2018.

\bibitem{Wang2019_2}
Jingyuan Wang, Yukun Cao, Ye~Du, and Li~Li.
\newblock {DST}: A deep urban traffic flow prediction framework based on
  spatial-temporal features.
\newblock In {\em Knowledge Science, Engineering and Management}, pages
  417--427. Springer International Publishing, 2019.

\bibitem{Cheng2017}
Anyu Cheng, Xiao Jiang, Yongfu Li, Chao Zhang, and Hao Zhu.
\newblock Multiple sources and multiple measures based traffic flow prediction
  using the chaos theory and support vector regression method.
\newblock {\em Physica A: Statistical Mechanics and its Applications},
  466:422--434, January 2017.

\bibitem{Sun2015}
Yuxing Sun, Biao Leng, and Wei Guan.
\newblock A novel wavelet-{SVM} short-time passenger flow prediction in beijing
  subway system.
\newblock {\em Neurocomputing}, 166:109--121, October 2015.

\bibitem{Xiao2018}
Jianli Xiao, Chao Wei, and Yuncai Liu.
\newblock Speed estimation of traffic flow using multiple kernel support vector
  regression.
\newblock {\em Physica A: Statistical Mechanics and its Applications},
  509:989--997, November 2018.

\bibitem{Polson2017}
Nicholas~G. Polson and Vadim~O. Sokolov.
\newblock Deep learning for short-term traffic flow prediction.
\newblock {\em Transportation Research Part C: Emerging Technologies},
  79:1--17, June 2017.

\bibitem{Wu2018}
Yuankai Wu, Huachun Tan, Lingqiao Qin, Bin Ran, and Zhuxi Jiang.
\newblock A hybrid deep learning based traffic flow prediction method and its
  understanding.
\newblock {\em Transportation Research Part C: Emerging Technologies},
  90:166--180, May 2018.

\bibitem{Lv2014}
Yisheng Lv, Yanjie Duan, Wenwen Kang, Zhengxi Li, and Fei-Yue Wang.
\newblock Traffic flow prediction with big data: A deep learning approach.
\newblock {\em {IEEE} Transactions on Intelligent Transportation Systems},
  pages 1--9, 2014.

\bibitem{Fu2016}
Rui Fu, Zuo Zhang, and Li~Li.
\newblock Using {LSTM} and {GRU} neural network methods for traffic flow
  prediction.
\newblock In {\em 2016 31st Youth Academic Annual Conference of Chinese
  Association of Automation ({YAC})}. {IEEE}, November 2016.

\bibitem{Yang2019}
Bailin Yang, Shulin Sun, Jianyuan Li, Xianxuan Lin, and Yan Tian.
\newblock Traffic flow prediction using {LSTM} with feature enhancement.
\newblock {\em Neurocomputing}, 332:320--327, March 2019.

\bibitem{Tian2018}
Yan Tian, Kaili Zhang, Jianyuan Li, Xianxuan Lin, and Bailin Yang.
\newblock {LSTM}-based traffic flow prediction with missing data.
\newblock {\em Neurocomputing}, 318:297--305, November 2018.

\bibitem{Wang2019}
Kang Wang, Kenli Li, Liqian Zhou, Yikun Hu, Zhongyao Cheng, Jing Liu, and Cen
  Chen.
\newblock Multiple convolutional neural networks for multivariate time series
  prediction.
\newblock {\em Neurocomputing}, May 2019.

\bibitem{Hochreiter1997}
Sepp Hochreiter and J\"{u}rgen Schmidhuber.
\newblock Long short-term memory.
\newblock {\em Neural Computation}, 9(8):1735--1780, November 1997.

\bibitem{Schmidhuber2015DeepLI}
J{\"u}rgen Schmidhuber.
\newblock Deep learning in neural networks: An overview.
\newblock {\em Neural networks : the official journal of the International
  Neural Network Society}, 61:85--117, 2015.

\bibitem{Kingma2014AutoEncodingVB}
Diederik~P. Kingma and Max Welling.
\newblock Auto-encoding variational bayes.
\newblock {\em CoRR}, abs/1312.6114, 2014.

\bibitem{google}
Google colab.

\bibitem{tensorflow}
Tensorflow.

\end{thebibliography}

\end{document}